%% file: acl_latex.tex
\pdfoutput=1

\documentclass[11pt]{article}

\usepackage[preprint]{acl}
\usepackage{amsmath, amssymb, amsfonts}
\usepackage{mathabx}

\usepackage{times}
\usepackage{latexsym}

\usepackage[T1]{fontenc}

\usepackage[utf8]{inputenc}

\usepackage{microtype}
\usepackage{booktabs}

\usepackage{inconsolata}

\usepackage{graphicx}

\title{How Do Language Models Hallucinate Legal Analysis, \\and How Can We Detect Them?}
\title{Analyzing Gaps of LLM-Generated Legal Analysis}
\title{Gazing Into the Gaps of Machine-Generated Legal Analysis}
\title{Gazing into the Gaps between 
Machine-Generated \\
and Human-Written Legal Analysis}
\title{Gaps or Hallucinations? Gazing into 
Machine-Generated \\
Legal Analysis for Fine-grained Text Evaluations}

\author{Abe Bohan Hou \textsuperscript{$\clubsuit$} \quad
  William Jurayj \textsuperscript{$\clubsuit$} \quad Nils Holzenberger \textsuperscript{$\spadesuit$}\\
  \textbf{Andrew Blair-Stanek} \textsuperscript{$\clubsuit$} \textsuperscript{$\diamondsuit$} \quad \textbf{Benjamin Van Durme} \textsuperscript{$\clubsuit$}\\
\textsuperscript{$\clubsuit$} Johns Hopkins University \textsuperscript{$\diamondsuit$} University of Maryland, Carey School of Law
\\
\textsuperscript{$\spadesuit$}Télécom Paris, Institut Polytechnique de Paris \\
  \texttt{bhou4@jhu.edu} \\
}

\newcommand{\clercs}{\textsc{Clerc\ }}

\definecolor{lightgreen}{RGB}{65,152,10}

\begin{document}
\maketitle
\begin{abstract}
Large Language Models (LLMs) show promise as a writing aid for professionals performing legal analyses.  However, LLMs can often hallucinate in this setting, in ways difficult to recognize by non-professionals and existing text evaluation metrics. In this work, we pose the question: \textbf{when can machine-generated legal analysis be evaluated as acceptable?} We introduce the neutral notion of \textit{gaps} -- as opposed to hallucinations in a strict erroneous sense -- to refer to the difference between human-written and machine-generated legal analysis.  Gaps do not always equate to invalid generation. Working with legal experts, we consider the \clercs generation task proposed in \citet{abe2024clerc}, leading to a taxonomy, a fine-grained detector for predicting gap categories, and an annotated dataset for automatic evaluation. Our best detector achieves 67\% $F_1$ score and 80\% precision on the test set. Employing this detector as an automated metric on legal analysis generated by SOTA LLMs, we find  around 80\% contain hallucinations of different kinds.\footnote{We release the code and data at \texttt{\url{https://github.com/bohanhou14/GapHalu}}.}

\end{abstract}
\begin{figure}
    \vspace{-1.28mm}
    \centering
    \includegraphics[width=\linewidth]{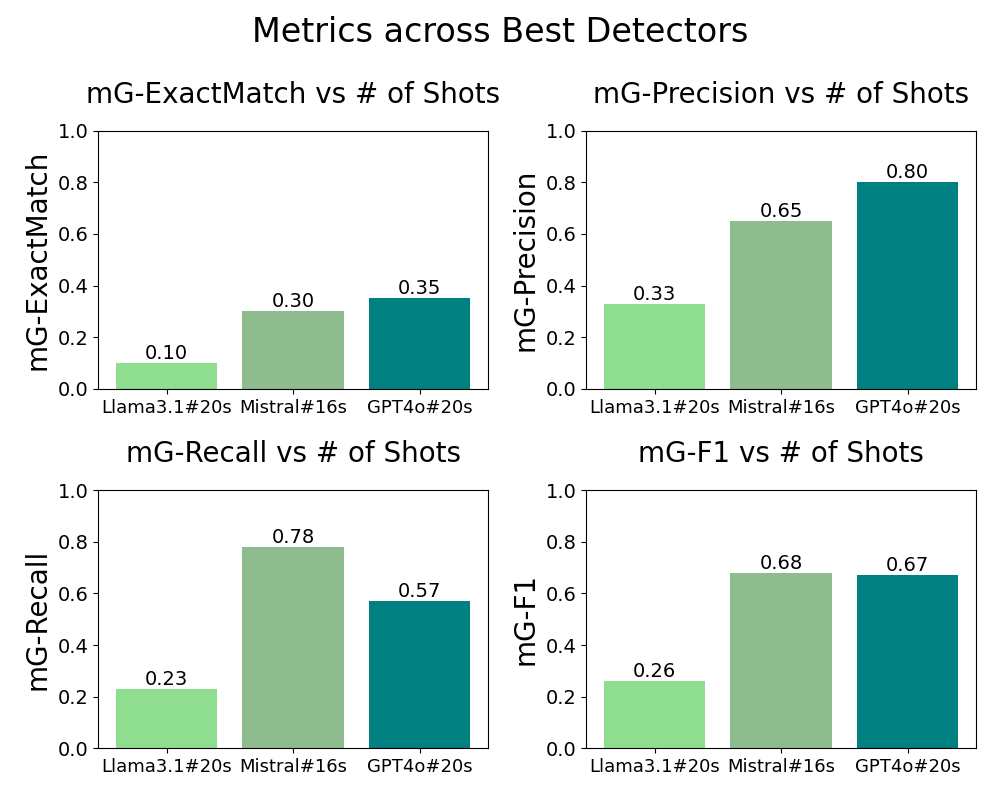}
    \caption{Detection results among the best detectors with different base models. $M\#ds$ means the best detector of base model $M$ has $d$ in-context demonstrations. \textbf{GPT-4o\#20s achieves the highest $mGEM$ and $mGP$, while Mistral-Nemo-Instruct-2407\#16s achieves the highest $mGR$ and $mGF_1$.}}
    \label{fig:main-res}
    \vspace{-3mm}
\end{figure}

\begin{figure*}
    \centering
    \includegraphics[width=\textwidth]{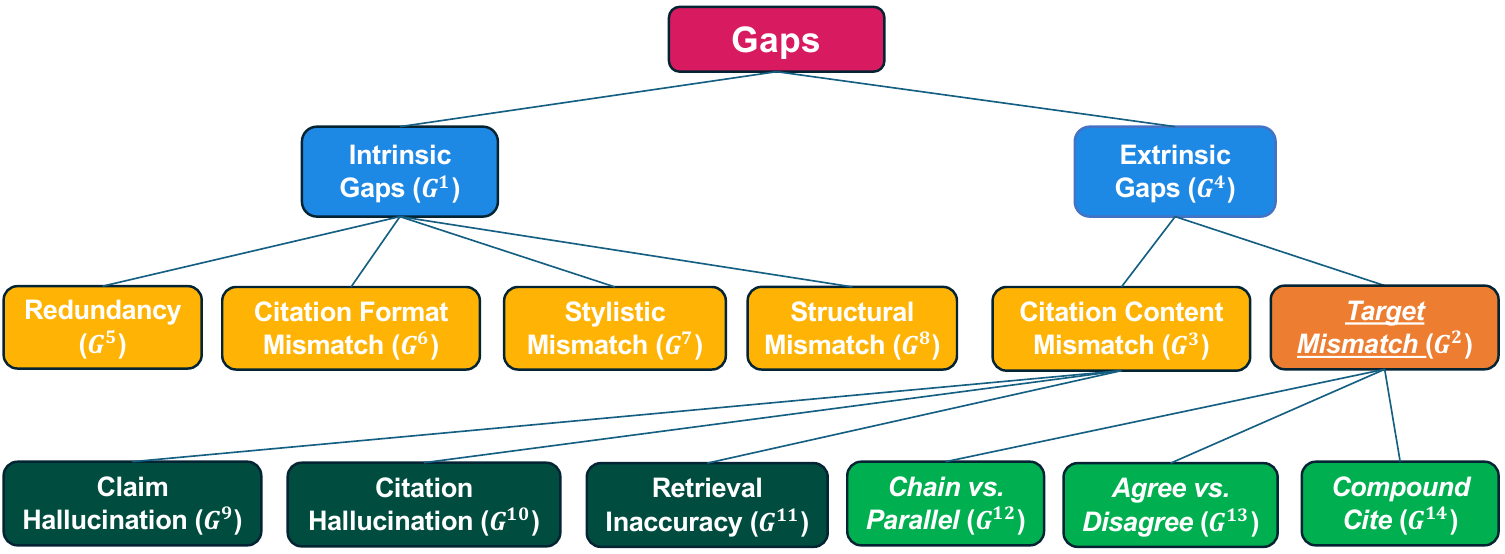}
    \caption{Our proposed taxonomy of gaps. Each category is discussed in depth in Section \ref{sec:taxonomy}. We highlight \textcolor{orange}{\textit{\underline{Target mismatch}}} ($G^2$) and its \textcolor{lightgreen}{\textit{child nodes}} ($G^{12}, G^{13}, G^{14}$) as we show they do not indicate hallucinations as opposed to other gap categories (with examples in Appendix \ref{app:examples}). Meanwhile, \textbf{citation content mismatch} and \textbf{intrinsic gaps} are generally considered hallucination and both indicate invalidity of generation.} 
    \label{fig:taxonomy-tree}
\vspace{-5mm}
\end{figure*}
\section{Introduction}
Legal professionals write legal analysis to help precisely communicate a legal issue or persuade judges \citep{cornell2024legalanalysis}. Despite recent work demonstrating that LLMs have the potential to generate realistic legal analyses to aid lawyers, they severely hallucinate \citep{abe2024clerc, Magesh2024HallucinationFreeAT}. In order to drive improvements, it is important to develop insights on the nature, categories, and sources of these hallucinations.

Evaluating legal analysis generation is challenging because the generation may: (1) have \textbf{multiple ground truths}, as legal practitioners can write an acceptable piece of analysis in many ways, (2) have \textbf{implicit and complex criteria to be judged based on legal expertise}, which makes obtaining human annotation data costly, (3) process \textbf{long-context}, which creates difficulties for evaluating faithfulness to previous contexts, and (4) involve \textbf{retrieving cited sources} and might propagate retrieval inaccuracies to downstream generation. A similar task to this is the automatic generation of research ideas, which is also challenging and expensive to evaluate \citep{Si2024CanLG, Lu2024TheAS}.  Evaluation of legal analysis is further complicated due to (5) \textbf{disagreement on the analysis and interpretation of law}, even between the most experienced legal professionals like Supreme Court judges. A law is interpreted both \textit{objectively} according to varying theories of legal interpretations, and also \textit{subjectively} according to the stance of the interpreter \citep{sep-legal-interpretation}. This is exemplified by the range of concurring and dissenting opinions written in the U.S. Supreme Court's 2022 \textit{Dobbs v. Jackson} decision overturning \textit{Roe v. Wade} \citep{Kaveny2023AbortionAT}.

\citet{abe2024clerc} propose a legal analysis task, evaluating the capabilities of models in generating an analytical paragraph, compared with the human-written paragraph (\textit{target}) from an original case using reference-based metrics such as ROUGE and BARTScore \citep{Lin2004ROUGEAP, Yuan2021BARTScoreEG}. Following common practice in evaluating generation systems, this evaluation scheme assumes the target is the ground truth and scores paragraphs less similar to the target as having lower validity and quality. In contrast, here we argue that \textbf{dissimilarities between machine-generated and human-written legal analyses are not necessarily errors or hallucinations}. We denote such dissimilarities with a neutral term, \textit{gaps}, inspired by the naming of \citet{pillutla-etal:mauve:neurips2021}, to show that dissimilarities are not determinant for evaluating generated analyses.

We also note that the general notion of gaps is not specific to legal analysis generation, but applicable to
any generation setting when multiple ground truths are possible. We focus on gaps in legal analysis generation as it is a domain especially suitable for this exploration. Unlike multiple translations or abstract summarizations which may have differences in syntax and word choice or the facts deemed essential to carry into a summary, valid legal analyses can illustrate significantly higher variability. Moreover, the creation of multiple references for legal analysis is cost prohibitive, as it requires legal experts to create an alternative writing that leads to the same result.

In this work, our contributions include:
\begin{enumerate}
    \item A detailed taxonomy of gaps to enable  more fine-grained evaluation of legal analysis.
    \item A manually annotated detection dataset, obtained by working with legal experts.
    \item  LLM-based detectors with best performance of 67\% $F_1$ and 80\% precision on the test set.  
    \item Automated evaluation metrics for legal analysis, \textsc{GapScore} and \textsc{GapHalu}, which reveal that around 80\% of \clercs generations using GPT-4o \citep{gpt424} and Llama-3-8B-Instruct \citep{Dubey2024TheL3} contain hallucinations.
\end{enumerate}

The rest of the work is laid out as follows: we provide background on legal analysis and hallucination in Section \ref{sec:background}, then explain our proposed gap taxonomy in Section \ref{sec:taxonomy}. We  develop detectors for classifying gaps according to the taxonomy in Section \ref{sec:gap_detect_overview} and apply these to evaluate legal analysis generations in Section \ref{sec:evaluate_legal_gen}. Lastly, we provide suggestions for mitigating legal hallucinations in Section \ref{sec:mitigation} and discuss related work in Section \ref{sec:related_work}.

\section{Background}
\label{sec:background}
\subsection{Legal Analysis Generation}
We create our taxonomy based on the legal analysis generation task proposed in \clercs \cite{abe2024clerc}, which is formulated as a retrieval-augmented next-paragraph prediction problem. Given a legal case document, which cites a set of other legal cases, $R$, to support its writing, an autoregressive language model (LM) is asked to generate the next paragraph based on its prefix and the cited external documents, formally as: ${\hat{p}_t \sim P_{LM}(\cdot | p_1, \dots, p_{t-1}, R)}$, where ${\hat{p}_t \sim P_{LM}(\cdot | p_1, \dots, p_{t-1})}$ is a paragraph-level notation we introduce to refer to the sampling of the next paragraph given previous paragraphs. Note that under this formulation, \clercs also assumes $R$ is given, whereas in practice this is achieved via an imperfect legal document retrieval process that makes this task more challenging \citep{abe2024clerc, mahari2023LePaRD}.
The machine generation is evaluated by comparing $\hat{p}_t$ to $p_t$, the human-written target paragraph. Since each case document has thousands of words, \clercs instead selects the most salient paragraph within each cited document as the input, ensuring it can fit the context window of a LLM. \clercs also guarantees the target paragraphs to contain analysis via a heuristic selection process, for which we refer readers to \citet{abe2024clerc} for details.

\begin{figure*}
\vspace{-10mm}
\centering
\fbox{
\footnotesize
\begin{minipage}{\textwidth}
\textbf{Generation}:\\
\dots
In Butner v. United States, 440 U.S. 48, 55, the Supreme Court emphasized that property rights in the assets of a bankrupt's estate are determined by state law unless a federal interest dictates otherwise. In this case, we must determine the debtor's obligations under Puerto Rican law and assess the applicability of interest on the debt owed to Nelson Torres Ruiz.

Consistent with the principles outlined in 114 B.R. 326, the determination of whether the debtor has a legal or equitable interest in the obligation to return the \$3,000 option money necessitates an analysis grounded in applicable nonbankruptcy state law. Article 1061 of the P.R. Civil Code, 31 Laws of P.R.Ann., Section 3025, establishes that where a debtor fails to perform an obligation, the creditor is entitled to interest as indemnification for damages and losses incurred.

Debtor's claim of impossibility of performance is addressed under Article 1138 of the Civil Code of Puerto Rico, but as established earlier, such a defense is unwarranted when the impossibility is not due to legal or physical barriers (31 Laws of P.R.Ann., Section 3193). Given the absence of such barriers, the return of the \$3,000 is mandated.

Furthermore, in alignment with 117 B.R. 15, the initiation of bankruptcy proceedings does not alter the creditor’s right to claim interest at the legal rate. Should the contract fail to specify an interest rate, Article 1061 defaults to a legal interest rate of 6\% per annum.

Thus, Mr. Bonilla is obliged to reimburse the full amount of \$3,000 plus the applicable legal interest rate of 6\% from the expiration of the option contract until the filing of the bankruptcy petition.
\\\\
\textbf{citations needed to make:} ['440 U.S. 48', '114 B.R. 326', '117 B.R. 15']
\\\\
\textbf{Target}:
\\
See generally, 3 Collier On Bankruptcy, (15th ed.) paragraph 502.02 (1990). See also, Butner v. U.S., 440 U.S. 48, 99 S.Ct. 914, 59 L.Ed.2d 136 (1979); In re MacDonald, 114 B.R. 326 (D.Mass. 1990); In re Milford Common J.V. Trust, 117 B.R. 15 (Bkrtcy.Mass., 1990).
\\\\
\textbf{Label}: [2]
\\\\
\textbf{Explanation}: The generated text provides detailed context and elaboration for each citation, whereas the target text chain cites them without additional detail. This indicates a target mismatch.
\end{minipage}
}
\caption{An example generated legal analysis from \clercs \citep{abe2024clerc}, labeled with 2 (target mismatch) and given an explanation. See the full version of this example and prompts to LLM-based detectors in Figure \ref{fig:full_example2}, \ref{fig:prompts}.}
\label{fig:small_example}
\end{figure*}

\subsection{Hallucination}
Numerous recent works have characterized hallucination \cite{ji2024hallucinationsurvey, Mishra2024FinegrainedHD, Chen2024UnifiedHD, Zhang2023AlleviatingHO}, and our definition of hallucination also aligns with prior works. We define hallucination as a span of LM-generated natural language which is incoherent, unfaithful to the contexts, or contain inaccurate or irrelevant information. As discussed in \citet{Zhang2023AlleviatingHO}, hallucinations can arise from three sources: conflicts with prompts to the language model, previous contexts, or facts. We adapt the notions of intrinsic and extrinsic hallucinations from \citet{ji2024hallucinationsurvey}, classifying whether a hallucination is intrinsic or extrinsic based on the \textbf{sources of conflicts:} conflicts with the prompts and previous contexts cause \textbf{intrinsic hallucinations}, whereas conflicts with external sources and facts induce \textbf{extrinsic hallucinations}.

\subsection{Hallucination in Legal Generation}
While there are various works dedicated to the understanding and mitigation of hallucinations in general \citep{Dhuliawala2023ChainofVerificationRH, Li2023HaluEvalAL, McKenna2023SourcesOH}, few have studied hallucinations in the legal domain \citep{Magesh2024HallucinationFreeAT, Dahl2024LargeLF}. \citet{Magesh2024HallucinationFreeAT} characterizes retrieval-augmented legal hallucinations based on two key criteria: \textbf{correctness}, which is whether the facts in the generation are correct and relevant to the prompt, and \textbf{groundedness}, which is whether the generation makes valid references to relevant legal documents. They also discuss a typology of retrieval-augmented generation errors consisting of four categories and analyze the contributing causes of the errors. In this work, we further breakdown the key criteria for determining hallucination, proposing a more fine-grained taxonomy consisting of 14 categories and introducing the notion of false positive hallucination (i.e. target mismatch). We also analyze the application of the taxonomy as automated evaluation metrics (\textsc{GapScore} and \textsc{GapHalu}) for legal analysis generation.

\section{A Taxonomy of Gaps}
\label{sec:taxonomy}
As popular reference-based metrics such as ROUGE and BARTScore \citep{Lin2004ROUGEAP, Yuan2021BARTScoreEG} and factuality metrics like FActScore \citep{Min2023FActScoreFA} only partially indicate validity of legal analysis generation
\citep{abe2024lfresco, abe2024clerc}, an automated metric for evaluating legal analysis generation is necessary. We first study the nature and typology of hallucinations, motivating a detailed taxonomy and error analysis, and then apply it to enable text evaluations (see Section \ref{sec:evaluate_legal_gen}).

We systematically review generation data from \clercs  and propose a detailed taxonomy of gaps in Figure \ref{fig:taxonomy-tree}. We classify the gaps into two types, in line with \citet{ji2024hallucinationsurvey}: \textit{intrinsic}, which refer to gaps that derive from the internal inaccuracies of LLMs in following prompts and previous contexts; and \textit{extrinsic}, which refer to gaps due to mismatches with cited sources and lacks of grounding on logical rules and existing facts. We attach examples in Figure \ref{fig:small_example} and Appendix \ref{app:full_example_and_prompts}, and for each fine-grained gap category in Appendix \ref{app:examples}.

\subsection{Intrinsic Gaps}
We discover and discuss four types of intrinsic gaps. \textbf{Redundancy} is when the generation appears to make repetitive statements (such as exact n-gram matches) and does not add further information to the analysis.
\textbf{Citation format mismatch} is when the generation appears not to match the standard styles of the uniform legal citation guide for US law, the Bluebook \citep{bluebook}, since \clercs is a US-specific legal dataset. Applying the taxonomy in international contexts, this gap can be adapted to the citation guides in other legal systems. \textbf{Stylistic mismatch} is when the generation uses an informal register or style of language that does not match with legalese. \textbf{Structural mismatch} is when the generation appears to generate the document from scratch or concludes the document prematurely, such as containing words like ORDER that typically appears at the beginning of case document, rather than predicting the next paragraph.

\subsection{Extrinsic Gaps}
We subdivide extrinsic gaps into two types.
\textbf{Target mismatch} refers to when the generation is obviously dissimilar from the target paragraph, but it can still be considered as another form of acceptable analysis.
\textbf{Citation content mismatch} refers to when the generation does not faithfully and factually reflect the content of the cited cases or hallucinate citations. We will discuss each subcategory in detail in this section.
 
\subsubsection{Target Mismatch}
We define three kinds of target mismatches, which are all caused by how the generation organizes the citations and their associated claims differently from the target. \textbf{Chain-versus-parallel} is when the target cites cases in a series (chain), all supporting the same claim, yet the generation elaborates every cited case and provides each with a claim. We also count the opposite scenario (i.e. the target does parallel and generation does chain) into this category. This gap is not necessarily unacceptable, as long as it does not make additional false claims, since it conveys the same overall meaning either in a concise or elaborate way. Similarly, \textbf{agree-versus-disagree} arises from mismatches on ways to characterize the relationship between multiple cited cases. The target might cite case A reversing the ruling in case B, whereas the generation might discuss case A and B respectively without highlighting the reversal relationship. \textbf{Compound cite} happens when the target combines the respective law from case A and B and makes a compound statement in a deductive manner, while the generation discusses them separately.

\subsubsection{Citation Content Mismatch}
We also discuss three kinds of citation content mismatches. \textbf{Claim hallucination} is when the claim supported by the citation is not truthful, not related to the context, or incoherent from cited paragraphs or the previous context. This was also discussed in \citet{abe2024lfresco} as the major hallucination scenario. Furthermore, we also have hallucinations caused by \textbf{retrieval inaccuracy}. Since the generated analysis needs to find external case documents as support, the retrieval process for documents can be inaccurate. To fit in the input context, the most salient chunk rather than the full text can be chosen, whose selection process might introduce additional inaccuracies. Lastly, \textbf{citation hallucination} refers to when the generated analysis contains non-existent citations, includes ones that were not supposed to appear, or omits citations that are supposed to be cited.

\subsection{When Are Legal Analyses Unacceptable?}

The presence of intrinsic gaps is generally considered intrinsic hallucinations, as they signal the failure of language models in understanding the task, following prompts and previous contexts, making coherent generations, and adapting linguistic styles appropriate to legal analyses. Among extrinsic gaps, citation content mismatch also qualifies as hallucination, for they all either introduce inaccurate information or contradict with the cited sources, in line with prior work on defining hallucinations \citep{Mishra2024FinegrainedHD, ji2024hallucinationsurvey}. On the other hand, we should not consider target mismatches as necessarily wrong since they mainly organize the information in a different way from the target paragraph (see examples in Figure \ref{fig:chain_cite}, \ref{fig:reverse_cite}, \ref{fig:compound_cite}). As legal analysis does not have a single definitive ground truth, the presence of target mismatch alone cannot indicate generation validity. 

We observe that generated analysis tends to include more than one category of gaps. Since intrinsic gaps and citation content mismatch are considered hallucinations in a stricter sense, we categorize generations that include any of them as unacceptable. On the contrary, if a generation does not include any of the gaps or \underline{only} includes target mismatch, we count it as acceptable.

\begin{table*}
\centering
\small
\begin{tabular}[\textwidth]
{@{}lllr@{}}
\toprule
Gap Name & Definition & Train Dist. (\%) & Test Dist. (\%) \\ \midrule
Intrinsic gaps (1)            & contradict with the instructions or context           & 13.79      & 18.18      \\
Target mismatch (2)           & organize info in a different way from the target      & 37.93      & 36.36      \\
Citation content mismatch (3) & contradict with the cited sources                     & 31.03      & 45.45      \\
No gaps (0)                   & more or less equivalent with the target paragraph     & 17.24      & 0.00        \\ \midrule
Dataset Balance & & \textbf{0.94} & \textbf{0.75} \\
\bottomrule
\end{tabular}
\caption{Detection dataset statistics. \textbf{Train/Test Dist.} refers to the distribution of labels in the train/test set. \textbf{Dataset Balance} is measured on a scale of [0-1], by the ratio of entropy of dataset labels over the entropy of perfectly balanced labels. A dataset is more balanced if the ratio is closer to 1, and \textbf{our dataset has a high balance.}}
\label{tab:dataset}
\vspace{-3mm}
\end{table*}
\begin{figure}
    \centering
    \includegraphics[width=\linewidth]{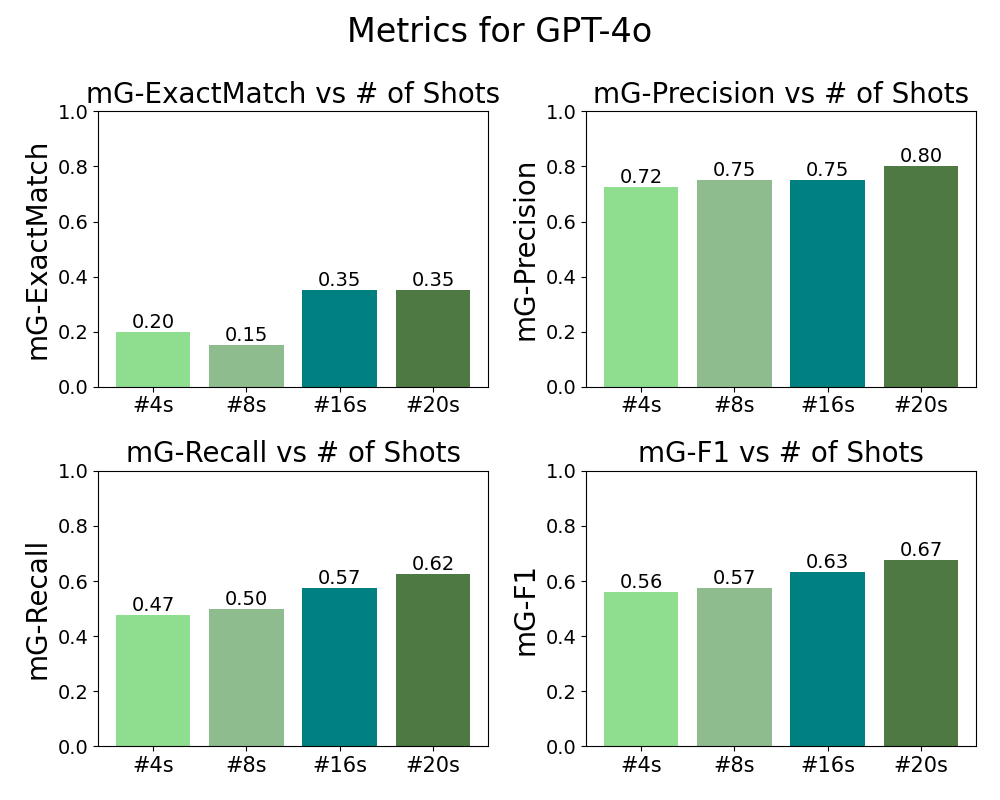}
    \caption{Detection results of the GPT-4o detector with different number of in-context demonstrations. \textbf{The full 20-shot detector yields the best overall detection accuracy}, while 16-shot has a marginal drop in accuracy.}
    \label{fig:abla-num-shots}
\end{figure}
\begin{figure}
    \centering
    \includegraphics[width=\linewidth]{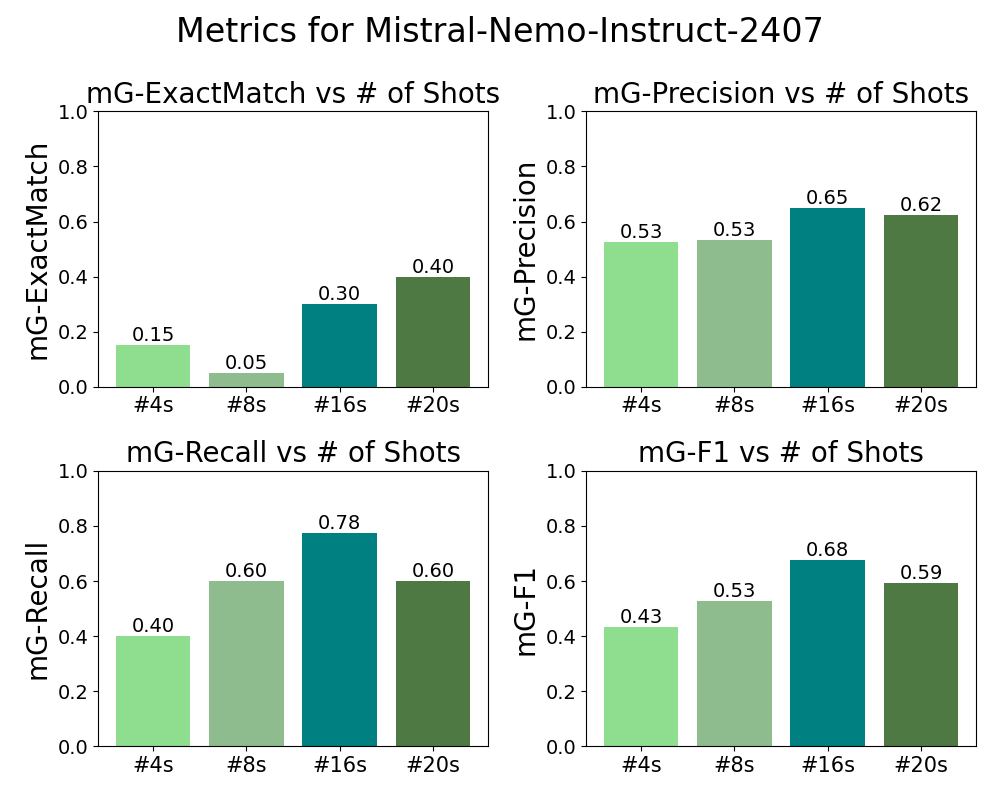}
    \caption{Detection results of the Mistral-Nemo detector with different number of in-context demonstrations. \textbf{The model achieves the maximal performance at 16 demonstrations and overfits at 20 demonstrations.}}
    \label{fig:mistral-abla-num-shots}
\end{figure}

\section{Gap Detection}
\label{sec:gap_detect_overview}
In this section, we build a detector to classify gap categories according to our proposed taxonomy. 
\subsection{Problem Formulation}
\label{sec:formulation}
Suppose we have $m$ total gap categories, given a piece of generated legal analysis $p_t$ which has gap categories ${g_t = (G^{1}_t, \dots, G^{m}_t)^\intercal}$, we predict the gap categories from a detector ${\hat{g}_t \sim f(\cdot | p_t)}$, where ${\hat{g}_t  = (\hat{G}^1_t, \dots, \hat{G}^{m}_t)^\intercal}$. ${f: D \rightarrow \mathbb{R}^{m}}$ is a detector function returning a $m$-dimensional vector, where each entry corresponds to a gap category, and the $k$-th gap for $t$-th generation exists if ${G^k_t = 1}$ and 0 otherwise.
We evaluate the detector on an arbitrary $i$-th piece of legal analysis with:
\begin{align*}
     \textsc{Gap-ExactMatch} (GEM) &= \mathbb{I}[\hat{g}_i = g_i] \\
     \textsc{Gap-Precision} (GP) &= \frac{\hat{g}_i \cdot g_i}{\lVert{\hat{g}_i}\rVert^2} \\
     \textsc{Gap-Recall} (GR) &= \frac{\hat{g}_i \cdot g_i}{\lVert{g_i}\rVert^2} \\
     \textsc{Gap-F1} (GF_1) &= \frac{2 GP \cdot GR}{GP + GR}
\end{align*}
where $\mathbb{I}$ is the indicator function, $g_i$ records gap categories of the $i$-th piece of legal analysis, and $\lVert\cdot\rVert$ is the norm of a vector. We calculate the mean of each metric over $N$ examples (e. g. ${mGEM = \frac{1}{N}\sum_{i=1}^{N} GEM_i}$) to reflect the overall performance of the detector.

\subsection{Experimental Setup}
We obtain and prepare our dataset from \clercs test set generations\footnote{\texttt{\url{https://huggingface.co/datasets/jhu-clsp/CLERC}}}. Due to the extraordinary expenses in hiring enough legal professionals for classifying 10 most granular gap categories (see Figure \ref{fig:taxonomy-tree}) and having enough data for each category, we choose to work at the second level of granularity, labeling each example with one or more from $\{$intrinsic gaps ($G^1$), target mismatch ($G^2$), citation content mismatch ($G^3$),  no gaps ($G^0$)$\}$. Although we do not label the most specific 10 categories ($G^5$ - $G^{14}$), we include and explain them in the instructions to annotators, which help clarify second-level gaps that are based on these bottom-level categories.

Working with legal experts\footnote{A tenured law professor who also co-authors this paper.}, we manually label 40 example generations respectively by GPT-4o \citep{gpt424} and Llama-3-Instruct-8B \citep{Dubey2024TheL3} (instructions in Appendix \ref{app:annotation_instruction}). We select 20 examples with an equal ratio of both model generations as the train set of the detector and the remaining 20 examples as the test set. Our detection dataset statistics is in Table \ref{tab:dataset}.

Our detector is based on prompting a long-context LLM with in-context demonstrations of examples labeled by humans \citep{brown2020language, rag20}, then asking it to predict the labels of a new example (prompts in Appendix \ref{app:prompts}). 
For the base model of our detector, we use GPT-4o \citep{gpt424}, Llama-3.1-8B-instruct, and Mistral-Nemo-Instruct-2407 \citep{Jiang2023Mistral7}. Our models are deployed with \texttt{vLLM} \citep{Kwon2023EfficientMM} with 1 A100 for Llama-3.1-8B-instruct and 4 A100s for Mistral-Nemo-Instruct-2407 to support the 128K tokens context window. 

We first label 20 examples along with brief explanations for the reasoning process behind our labeling. The prompt for our detector includes a summary of the instructions for human annotators and at most 20 labeled examples as in-context demonstrations. We also conduct an ablation study varying the number of demonstrations and present the results in Figure \ref{fig:abla-num-shots}. To assess the detector accuracy, we prompt it to predict 20 unlabeled examples and then manually label them, evaluating the mean of metrics discussed in Section \ref{sec:formulation} respectively, namely $mGEM$, $mGP$, $mGR$, and $mGF_1$.  

\begin{figure}
    \centering
    \includegraphics[width=\linewidth]{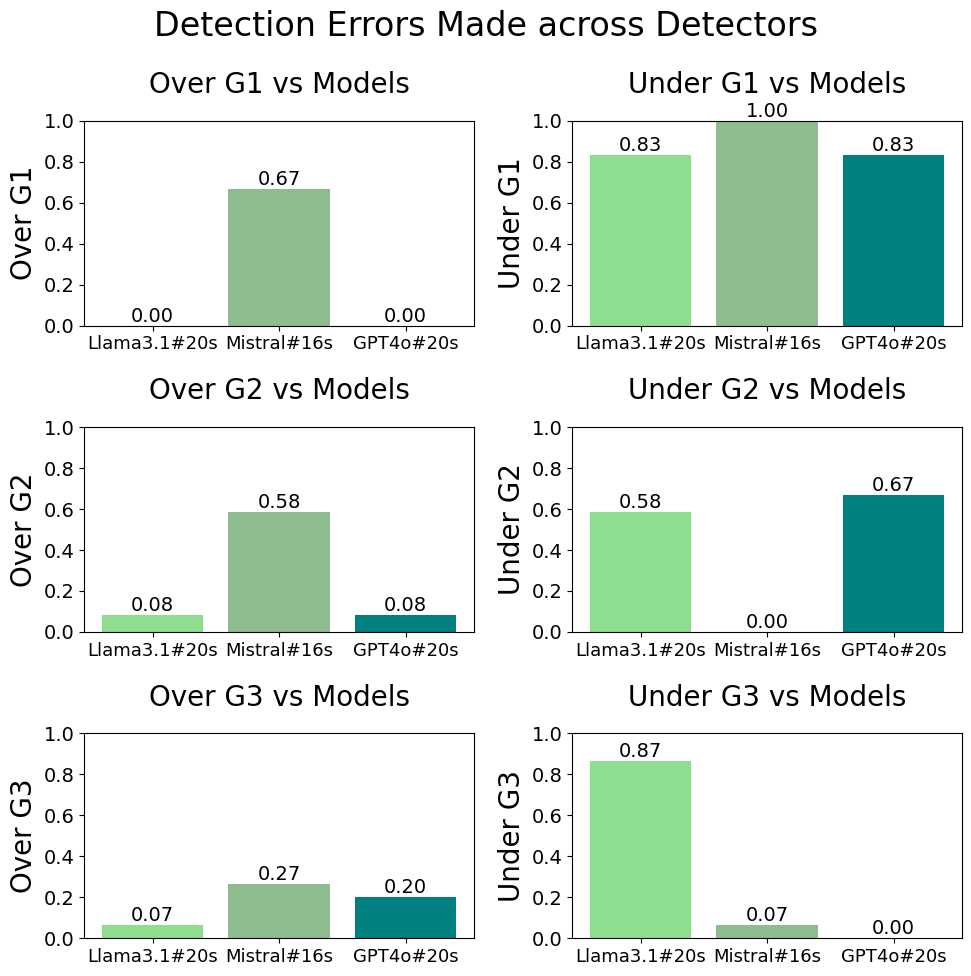}
    \caption{Error analysis of different detectors. \textbf{G1} refers to \textit{intrinsic gaps}, \textbf{G2} refers to \textit{target mismatch}, and \textbf{G3} refers to \textit{citation content mismatch}. \textbf{Mistral-Nemo tends to over-predict the presence of gaps, whereas GPT-4o and Llama-3.1 tend to under-predict.}}
    \label{fig:error_analyze_det}
    \vspace{-5mm}
\end{figure}
\subsection{Detection Results}
\label{sec:det_res}

Figure \ref{fig:main-res} compares the performances of three best detectors for each base model and discover that GPT-4o achieves the maximum $mGEM$ and $mGP$ with 20 demonstrations, while Mistral-Nemo-Instruct-2407 achieves the maximum $mGR$ and also $mGF_1$ with 16 demonstrations, by a small margin over GPT-4o, with the Llama-3.1-8B-instruct detector with 20 demonstrations being the worst among the three. We select the best detector for each base model according to our ablation studies on the number of in-context demonstrations. We find that the optimal number of in-context demonstrations is different for each model, with results presented in Figure \ref{fig:abla-num-shots}, \ref{fig:mistral-abla-num-shots}, \ref{fig:llama-abla-num-shots}.

To further understand the behavior and biases of detectors, we analyze the percentages of each label being over-predicted and under-predicted and present the results in Figure \ref{fig:error_analyze_det}. The Mistral-Nemo detector tends to over-predict across all gap categories, which explains why it has a high recall but relatively low precision compared to the GPT-4o detector. On the other hand, the GPT-4o detector under-predicts $G^1$ and $G^2$ but overall maintains the highest precision and exact match. Llama-3.1-Instruct has the worst performance. The three detectors all tend to under-predict $G^1$, which can be caused due to a relative lack of $G^1$ training data, or that the detection of $G^1$ is challenging per se. 

In sum, since $G^2$ generally does not indicate invalidity, \textbf{the GPT-4o detector is most useful to evaluate legal analysis generations as it is most accurate at identifying $G^1$ and $G^3$.}

\begin{table}[]
\centering
\small
\begin{tabular}{@{}llr@{}}
\toprule
Metric ($\times$ 100) & GPT-4o & Llama-3-8B-Instruct \\ \midrule
R1 $\uparrow$         & \textbf{26.73}  & 24.88               \\
R2 $\uparrow$         & \textbf{10.13}  & 8.86                \\
RL $\uparrow$         & \textbf{24.83}  & 23.20               \\
BF $\uparrow$         & \textbf{-3.13}  & -3.33               \\
\textsc{GapScore} $\downarrow$ & 96.31  & \textbf{95.46}               \\
\textsc{GapHalu} $\downarrow$  & \textbf{79.51}  & 82.05               \\ \midrule
$G^1$ $\downarrow$ &\textbf{24.80}&25.20 \\
$G^2$ $\downarrow$ &\textbf{82.99} &84.96 \\
$G^3$ $\downarrow$ &61.48 &\textbf{60.94} \\
\bottomrule
\end{tabular}
\caption{Evaluation of GPT-4o and Llama-3 \clercs generations with the Mistral-Nemo detector and F-Scores of ROUGE and BARTScore (BF). GPT-4o has higher \textsc{GapScore} while lower \textsc{GapHalu} than Llama-3, meaning that it has less hallucination. Over the fine-grained categories, GPT-4o has lower proportion of intrinsic gaps and target mismatches but higher percentage of citation content mismatch than Llama-3. \textbf{Both models generate legal analysis with severe hallucinations, as $\sim$ 80\% (indicated by \textsc{GapHalu}) contain hallucinations.}}
\label{tab:gs_gh_eval}
\vspace{-2mm}
\end{table}

\section{Re-Evaluate Legal Analysis Generation}
\label{sec:evaluate_legal_gen}
\subsection{\textsc{GapScore} and \textsc{GapHalu}}
In this section, we discuss an application of the detector in evaluating legal analysis generations. With a fine-grained detector, we can distinguish between generations with intrinsic gaps, target mismatches, and citation content mismatch, enabling the fine-grained evaluation of legal analysis genration. We propose the following metrics:
\begin{align*}
    \textsc{GapScore} &= \frac{1}{N}  \sum_{i=1}^{N}m_1(G^1_i + G^2_i + G^3_i) \\
    \textsc{GapHalu} &= \frac{1}{N} \sum_{i=1}^{N}m_1 (G^1_i + G^3_i)
\end{align*}
where $G^k_i$ is a binary variable that returns 1 when the $i$-th example contains $G^k$ and 0 otherwise.  $m_1(\cdot)$ refers to $\min(\cdot, 1)$. \textsc{GapScore} measures the ratio of $N$ examples having gaps, and \textsc{GapHalu} measures the ratio of hallucinations.
\subsection{Experimental Setup}
We sample 500 GPT-4o and Llama-3-8B-Instruct generations from \clercs respectively, and evaluate with the detector developed in Section \ref{sec:gap_detect_overview}. While GPT-4o detector has the highest accuracy at identifying hallucinations, we run the Mistral-Nemo detector due to significant expenses incurred in accessing the GPT-4o API. We also run ROUGE and BARTScore evaluations over the texts for a comparison with \textsc{GapScore} and \textsc{GapHalu}. 

\subsection{Re-Evaluation of Legal Analysis}
Table \ref{tab:gs_gh_eval} presents results of evaluating legal analysis generations with automated metrics. Our experimental results of ROUGE and BARTScore highly align with the results in \citet{abe2024clerc}. 

We discover that GPT-4o generations have less hallucination compared to Llama-3-8B-Instruct, as indicated by a lower \textsc{GapHalu} score. However, it has a slightly higher proportion of citation content mismatch ($G^3$). As our proposed taxonomy classifies citation hallucination as a type of citation content mismatch, this result is partially explained by the findings in \citet{abe2024clerc} that GPT-4o tends to hallucinate more false positive citations than other models. 

In addition, we find that Llama-3-8B-Instruct generations tend to have more target mismatch, which might explain why they score lower on ROUGE and BARTScore. Since target mismatch often features obvious dissimilarities (see examples in Figure \ref{fig:chain_cite}, \ref{fig:reverse_cite}), having a higher proportion of target mismatch potentially causes a great lack of textual overlap and lowers reference-based metrics like ROUGE and BARTScore more significantly. 

Overall, we discover that \textbf{around 80\% of the generated legal analysis contain hallucinations like intrinsic gaps and citation content mismatch, which indicates the limitation of SOTA LLMs at generating legal analyses.} We estimate the actual percentage of legal hallucinations to be even higher, as we discuss in Section \ref{sec:det_res} that the Mistral-Nemo detector tends to under-predict the presence of intrinsic gaps.

\section{Mitigation Suggestions}
\label{sec:mitigation}
In this section, we discuss general strategies to mitigate legal hallucinations as well as specific suggestions related to each gap category.
\subsection{General Strategies}
Intrinsic gaps often arise from failures to follow prompts and previous contexts, lack of adaptation to the linguistic styles and citation formats of the legal domain. Target mismatches also reflect that LLMs struggle with finding patterns consistent with human preferences to organize information in legal writing. Therefore, we suggest continued pre-training of SOTA LLMs on the legal domain with similar approaches in \citet{chalkidis2020legalBERT, Niklaus2024FLawNT5AE, Gururangan2020DontSP} to address the model domain shift and improve its adaptation to legalese.

Furthermore, \textbf{decomposition of the reasoning structure in legal analysis} may critically improve generation quality and mitigate hallucinations, and even improve retrieval of cited sources. A legal case is usually structured with an introduction and summary of facts, an identification of the core dispute, and then breaks down the core dispute into subclaims to be analyzed with, until an eventual logical conclusion is formed. The reasoning is hierarchical, which enables extraction of an explicit structure. Such reasoning structure can be utilized to enhance downstream applications via combining with prompting or with a symbolic solver \citep{weir2024enhancingsystematicdecompositionalnatural}. LLMs would be able to parse missing points from the reasoning structure and generate the necessary information, and avoid claims already addressed. A complex legal reasoning task can be effectively decomposed into simpler sub-problems, enabling the generation of high-quality legal analysis through a divide-and-conquer strategy.

\subsection{Intrinsic Gaps}
Aside from the general strategies, intrinsic gaps also indicate that LLMs may struggle with using the correct citation formats in legal writing. We suggest incorporating domain-specific knowledge about legal citations through fine-tuning, RAG, or tool use \citep{Dubey2024TheL3, Yang2023GPT4ToolsTL}.

\subsection{Extrinsic Gaps}
Extrinsic hallucinations in retrieval-augmented legal analysis generation can be attributed to conflicts with the cited sources or the cited sources retrieved being irrelevant. Improving retrieval architecture, especially with long-context retrieval strategy with awareness of the latent logical structure, can be one critical direction to improve generation and mitigate hallucinations \citep{sarthi2024raptor}.

\section{Related Work}
\label{sec:related_work}
\subsection{Citation Ontology}
Even before internet-scale citation graphs were tractable, bibliometric research focused on the social and cognitive implications of different citation schemata \cite{cronin_need_1981}. \citet{peroni_fabio_2012}'s popular framework categorizes citations based on the factual and rhetorical roles that the cited document plays in the citing paper. More recent work has used LLMs to generate or classify citations in scientific literature \cite{cohan-etal-2019-structural, xing_automatic_2020, luu_explaining_2021}. 

\subsection{Argument Analysis}
Generating and analyzing persuasive arguments is another useful formulation for case-based legal writing. Some efforts have explored how various argument rating approaches can train models to persuade more effectively \cite{mouchel2024logicalfallacyinformedframeworkargument, durmus2024persuasion}.  \citet{saha-etal-2021-explagraphs} use human annotations to train a system that converts textual arguments into logical graphs. By searching over these graphs, LMs can generate deductive arguments to prove or disprove claims based on evidence from cited documents \cite{Weir2022NELLIEAN, Sanders2024TVTREESME}.

\subsection{Legal Reasoning}
Legal reasoning is challenging even for the most powerful LMs \cite{stanek2023gpt3statutory}. Fine-tuning smaller LMs can result in higher performance over generic models \cite{Niklaus2024FLawNT5AE, chalkidis2020legalBERT}. An alternative appraoch is to integrate symbolic solvers during reasoning \cite{Padhye2024SoftwareEM, Holzenberger2023ConnectingSS}.

\section{Conclusion and Future Work}
To facilitate a fine-grained evaluation of generated legal analysis, we propose a taxonomy of gaps and develop detectors to analyze the sources of legal hallucinations, also experimenting with \textsc{GapScore} and \textsc{GapHalu} to assess the validity of generated legal analysis. For future work, we will extend our framework of analyzing gaps on the general text domain for fine-grained text evaluations.

\section*{Limitations}
Our work builds up the foundation for legal hallucination evaluation metrics, but the detection of gaps can be imperfect, since the LLMs used as the base models of the detectors generally struggle on legal tasks and experience domain shifts \citep{stanek2023gpt3statutory, blairstanek2024bltlargelanguagemodels, chalkidis2020legalBERT}. Moreover, the parsing of legal citations is still an open problem to the legal NLP community, and this imperfect process introduces minor inaccuracies that propagate to affect the robustness of our detectors. 

\section*{Ethical Considerations}
Our work concerns with U.S. historical law data, with cases dated earliest from the year of 1658 \citep{cap}. The data might express outdated views and ideologies, such as racism and sexism, which are disturbing and considered unethical to the current academic community. It raises interesting questions and needs for further discussions on how we can strike the balance between generating safe and harmless speech, versus having to process controversial laws and historical legal facts to produce accurate analyses.

\section*{Acknowledgement}

This work was supported in part by the U.S. National Science Foundation under grant 2204926. Opinions, findings, and conclusions or recommendations expressed in this article come from the authors and do not reflect the views of the National Science Foundation. We also thank JHU CLSP members Guanghui Qin and Orion Weller for their advice on data analysis and paper writing, Brian Lu for his feedback on the figure design, as well as Tianjian Li and Jingyu Zhang for suggestions on the paper organization.

\bibliography{custom}

\appendix

\section{A Full Example of Legal Analysis and Example Annotations}
\label{app:full_example_and_prompts}

We present the full version of the example from Figure \ref{fig:small_example} in Figure \ref{fig:full_example1} and \ref{fig:full_example2}. An example annotation is included after the line break in Figure \ref{fig:full_example2}.

\section{Examples of Fine-grained Gap Categories}
\label{app:examples}
We present examples of the most fine-grained gap categories ($G^5$ - $G^{14}$) in Figure \ref{fig:example_3},
\ref{fig:example_3_exp}
\ref{fig:example_11}, \ref{fig:example_11_exp}, and particularly, examples of target mismatches in Figure 
\ref{fig:chain_cite}, \ref{fig:reverse_cite}, \ref{fig:compound_cite}.

\section{Annotation Instruction}
\label{app:annotation_instruction}
We present the annotation instruction for human annotators in Figure \ref{fig:instructions-fig1} and \ref{fig:instructions-fig2}.

\section{Prompts to LLM Detectors}
\label{app:prompts}
We present prompts to LLM detectors in Figure \ref{fig:prompts}. The \texttt{variables} in \{\} are specific inputs to the prompt, and we vary $k$ demonstraations for conducting the ablation studies.

\input{examples/full_example}

\begin{figure}
    \centering
    \includegraphics[width=\linewidth]{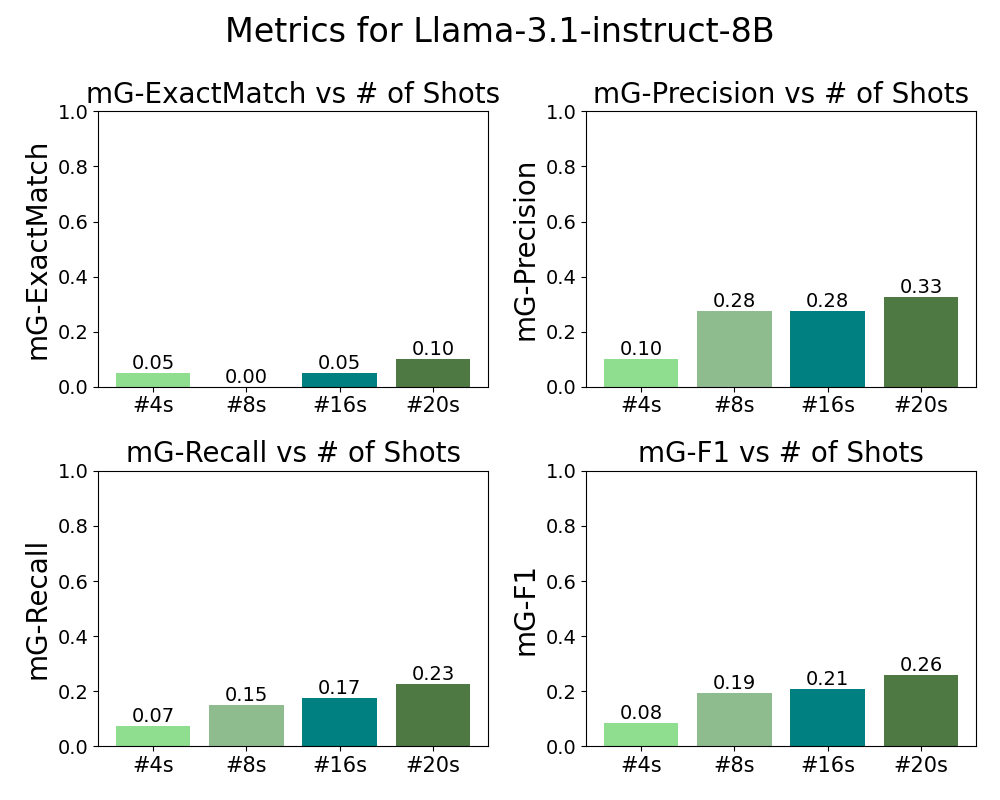}
    \caption{Detection results of the Llama-3.1-8B-instruct detector with different number of in-context demonstrations. \textbf{The model achieves the maximal performance at 20 demonstrations.}}
    \label{fig:llama-abla-num-shots}
\end{figure}

\input{examples/example_3}

\input{examples/example_11}

\input{examples/target_mismatch_examples}
\input{examples/annotation_instructions}
\input{examples/prompts_to_detectors}

\end{document}

%% file: examples/full_example.tex
\begin{figure*}[t]
\centering
\fbox{
\footnotesize
\begin{minipage}{\textwidth}
\textbf{Generation}:\\
\dots
In Butner v. United States, 440 U.S. 48, 55, the Supreme Court emphasized that property rights in the assets of a bankrupt's estate are determined by state law unless a federal interest dictates otherwise. In this case, we must determine the debtor's obligations under Puerto Rican law and assess the applicability of interest on the debt owed to Nelson Torres Ruiz.

Consistent with the principles outlined in 114 B.R. 326, the determination of whether the debtor has a legal or equitable interest in the obligation to return the \$3,000 option money necessitates an analysis grounded in applicable nonbankruptcy state law. Article 1061 of the P.R. Civil Code, 31 Laws of P.R.Ann., Section 3025, establishes that where a debtor fails to perform an obligation, the creditor is entitled to interest as indemnification for damages and losses incurred.

Debtor's claim of impossibility of performance is addressed under Article 1138 of the Civil Code of Puerto Rico, but as established earlier, such a defense is unwarranted when the impossibility is not due to legal or physical barriers (31 Laws of P.R.Ann., Section 3193). Given the absence of such barriers, the return of the \$3,000 is mandated.

Furthermore, in alignment with 117 B.R. 15, the initiation of bankruptcy proceedings does not alter the creditor’s right to claim interest at the legal rate. Should the contract fail to specify an interest rate, Article 1061 defaults to a legal interest rate of 6\% per annum.

Thus, Mr. Bonilla is obliged to reimburse the full amount of \$3,000 plus the applicable legal interest rate of 6\% from the expiration of the option contract until the filing of the bankruptcy petition.
\\\\
\textbf{citations needed to make:} ['440 U.S. 48', '114 B.R. 326', '117 B.R. 15']
\\\\
\textbf{Target}:
\\
See generally, 3 Collier On Bankruptcy, (15th ed.) paragraph 502.02 (1990). See also, Butner v. U.S., 440 U.S. 48, 99 S.Ct. 914, 59 L.Ed.2d 136 (1979); In re MacDonald, 114 B.R. 326 (D.Mass. 1990); In re Milford Common J.V. Trust, 117 B.R. 15 (Bkrtcy.Mass., 1990).
\\

\textbf{reference\_case\_1:} 440 U.S. 48

506 F. 2d 1242, 1243 (CA4 1974). See generally 4A W. Collier, Bankruptcy ¶ 70.16, pp. 157-165 (14th ed. 1975); Hill, The Erie Doctrine in Bankruptcy, 66 Harv. L. Rev. 1013 (1953). In some title States, the mortgagee’s right to rents and profits may be exercised even prior to default, see Me. Rev. Stat. Ann., Tit. 33, § 502 (1964); in all events, the right at least attaches upon default, see Uvalda Naval Stores Co. v. Cullen, 165 Ga. 115, 117, 139 S. E. 810, 811 (1927). See generally R. Kratovil, Modern Mortgage Law and Practice § 294, p. 204 (1972); Comment, The Mortgagee’s Right to Rents and Profits Following a Petition in Bankruptcy, 60 Iowa L. Rev. 1388, 1390-1391 (1975). North Carolina has been classified as a “title” State, Comment, The Mortgagee’s Right to Rents After Default, 50 Yale L. J. 1424, 1425 n. 6 (1941), although it does not adhere to this theory in its purest form. Under its case law, a mortgagee is entitled to possession of the mortgaged property upon default, and need not await actual foreclosure. Such possession might be secured either with the consent of the mortgagor or by an action in ejectment. But so long as the mortgagor does remain in possession, even after default, he — not the mortgagee — appears to be entitled to the rents and profits. See Brannock v. Fletcher, 271 N. C. 65, 155 S. E. 2d 532 (1967); Gregg v. Williamson, 246 N. C. 356, 98 S.
\\

\textbf{reference\_case\_2:} 114 B.R. 326

U.S.C. § 363(b)(1). “Property of the estate” includes “all legal or equitable interests of the debtor in property as of the commencement of the case.” 11 U.S.C. § 541(a)(1). It is “necessary to look to nonbankruptcy law, usually state law, to determine whether the debtor has any legal or equitable interest in any particular item.” 4 Collier on Bankruptcy, ¶ 541.02[1] at 541-13 (15th ed. 1989). Since “property interests are created and defined by state law,” such interests are analyzed under state law in bankruptcy proceedings unless “some federal interest requires a different result.” Butner v. United States, 440 U.S. 48, 55, 99 S.Ct. 914, 918, 59 L.Ed.2d 136 (1979). See also In re Prichard Plaza Associates Ltd. Partnership, 84 B.R. 289, 293 (Bankr.D.Mass.1988). This Court’s resolution of the dispute over the debtor’s interest in Spectrum Wire is grounded in state corporations law, and takes into account the equitable powers of the bankruptcy court. However, in light of the bankruptcy court decision under review, this Court’s analysis of the appellants’ ownership interest in the Spectrum Wire stock must begin with a discussion of the Massachusetts law of trusts. A. Stock as the Subject Matter of an Express Trust The bankruptcy court found that conduct and verbal agreements by the debtor’s father “manifested an intention to hold in trust for the Debtor the shares of Spectrum stock standing in the father’s name.” In re MacDonald, 101 B.R. at 841. This conclusion, that an express trust was created by the debtor’s father,
\\

\textbf{reference\_case\_3:} 117 B.R. 15

order against the debtor. The automatic stay prevented any further action by the Bank, including service of the restraining order. The debtor has remained in physical possession and has continued to collect all of its rents. The Bank promptly filed with the bankruptcy court an emergency motion for relief from stay and for authority to continue with its possession and to collect the rents. The law was clarified by the United States Supreme Court in 1979 in the case of Butner v. United States, 440 U.S. 48, 99 S.Ct. 914, 59 L.Ed.2d 136, 4 B.C.D. 1259. The court held that: ... Congress has generally left the determination of property rights in the assets of a bankrupt’s estate to state law. Property interests are created and defined by state law. Unless some federal interest requires a different result, there is no reason why such interests should be analyzed differently simply because an interested party is involved in a bankruptcy proceeding. Looking to Massachusetts law, an assignment of rent gives the mortgagee a valid security interest which becomes effective upon a default and an overt act by the mortgagee to take actual or constructive possession. Bankruptcy does not change the assignee/mortgagee’s right to the rent so long as possession was obtained pre-fil-ing, or a request is made to the bankruptcy-court for relief. The matter was further extensively analyzed by Bankruptcy Judge James F. Queenan, Jr. in the case of In re Prichard Plaza Associates Limited Partnership, 84 B.R. 289 (Bkrtcy.D.Mass.1988). For a
\\

(CONTINUED NEXT PAGE)
\end{minipage}
}
\caption{(1/2) A full example machine-generated legal analysis from \citep{abe2024clerc}, with previous context and cited sources provided. Texts after the line break are added when prompting LLM-based detectors.}
\label{fig:full_example1}
\end{figure*}
\begin{figure*}[t]
\centering
\fbox{
\footnotesize
\begin{minipage}{\textwidth}
(CONTINUED FROM LAST PAGE)\\

\textbf{previous\_text:} 
\\
OPINION AND ORDER
SARA E. de JESUS, Bankruptcy Judge.
The matter pending before the Court is whether creditors Nelson and Elizabeth Torres are entitled to the payment of interest on Claim \# 13, and the applicable interest rate.
Pursuant to Debtor’s request for a valuation of claim \# 13, we held an evidentiary hearing. The parties have agreed to the following facts:
“a. That on July 22, 1980, Nelson Torres Ruiz and Adrián Bonilla Montalvo signed an Option Contract for the purchase of a plot of land marked number twenty (20).
b. The price of said plot of land was \$7,250.00, of which at the signing of the Option Contract, Nelson Torres Ruiz paid Adrián Bonilla Montalvo the sum of \$500.00 and later that same day paid him \$2,500.00 for a total of \$3,000.00.
c. The Option Contract enumerated a period of two years from the date of signing within which the debtor, Adrián Bonilla Montalvo, was to execute the purchase deed or reimburse Nelson Torres Ruiz the sum of \$3,000.00.,
d. That Mr. Nelson Torres Ruiz was single when he entered into an option agreement for certain lot of land on July 22, 1980.
e. That Mr. Nelson Torres Ruiz gave Mr. Adrián Bonilla \$3,000.00 as option money-
f. That debtor according to clause \# 6 of the option contract is obliged, and has accepted to do so, to return to this creditor the \$3,000.00.
g. That debtor has recognized the debt of \$3,000.00 owed to Mr. Nelson Torres and has scheduled the same as \$900.00 priority and \$2,100.00 as general unsecured claim.
h. Mrs. Elizabeth Hermida de Torres married Mr. Nelson Torres Ruiz after the option contract was signed.
i. Mrs. Elizabeth Hermida de Torres was not a party to the option contract signed on July 22, 1980 by debtor and Mr. Nelson Torres.
j. .That on January 12, 1984, Mrs. Elizabeth Hermida de Torres was deputy clerk of the Superior Court of Puerto Rico, Courtroom of Mayaguez.
k. That Banco Comercial de Mayaguez filed suit number 81-1138 against debtor and his ex-wife on the Superior Court of Puerto Rico, Courtroom of Mayaguez.
l. That on June 7, 1983 Attorney Jovino Martinez wrote a letter to debtor on behalf of Mr. Nelson Torres requesting the return of the option money given by him to debt- or.
m. The plot of land where Mr. Nelson Torres had his option was sold after the filing for relief and with the authority of this Court.”
Two Joint Exhibits were also admitted: the Option contract executed by the Debtor and Nelson Torres on July 22, 1980; and a letter dated June 7, 1983 from Attorney Jovino Martinez Ramirez to Attorney Adrián Bonilla Montalvo requesting the return of the money paid by Mr. Torres plus legal interest.
CONCLUSIONS OF LAW
In bankruptcy, issues as to the validity and legality of a claim are determined pursuant to applicable state law. Thus, we must decide the question at hand by applying the pertinent Articles of the Civil Code of P.R.
The option contract executed by Debtor and Nelson Torres Ruiz, called for the execution of the deed of sale within two years from July 22, 1980. However, the contractual terms also required Mr. Bonilla to return the total price for the option, if he could not obtain the permits required by the local government allowing him to segregate and sell the optioned plot, within this same two year period. The contract does not mention interest payments.
The Debtor raises the defense of impossibility of compliance with the obligation in order to release himself from the obligation and/or any liability. Mr. Bonilla claims a legal and physical impossibility based on events which occured almost six years after the Option contract had expired, and, in any event, these events concern his fiscal or monetary problems. Article 1138 of the Civil Code of Puerto Rico, 31 Laws of P.R.Ann., Section 3193, provides that, “In obligations to do, the debtor shall also be released when the prestation appears to be legally or physically impossible.” However, Debtor’s reliance on this Article of the Civil Code is unwarranted inasmuch as the legal and physical impossibility contemplated by law are not present in this contested matter.
Article 1061 of the P.R.Civil Code, 31 Laws of P.R.Ann., Section 3025, provides that when the obligation consists in the payment of a sum of money, and the person incurs in default, the creditor is entitled to be indemnified for damages and losses suffered, which will consist in the payment of interest. If the parties failed to agree upon the payment of interest and or the interest rate, then the interest to be paid will be the legal interest at the applicable rate. Furthermore, “Until another rate is fixed by the Government, interest at the rate of six percent per annum shall be considered as legal.”
Under these circumstances, Mr. Bonilla must reimburse the full amount of the option contract paid by Mr. Torres, plus interest at the legal rate of 6\% per year, from the date the option contract expired to the date this petition was filed, pursuant to 11 U.S.C. Section 502(b)(2).
. During the hearing, Nelson Torres Ruiz and his wife Elizabeth waived their claims for damages other than interest discussed herein.
. 11 U.S.C. Section 502(c)(2).
. It was also undisputed that Debtor was not able to secure the government permits needed to segregate and sell the plot to the Optionee, and the deed of sale was never executed.
. The Court takes judicial notice that the petition for bankruptcy pursuant to Chapter 11 of the Code was entered on February 19, 1987; and that Nelson and Elizabeth Torres, filed proof of claim no. 13 in the amount of \$25,-000.00. Attached thereto are the copies of the following documents: the option contract entered into by the debtor Adrián Bonilla Montal-vo and Nelson Torres Ruiz on July 22, 1980; the complaint filed by Mr. Torres Ruiz against the debtor before the Superior Court of Puerto Rico, Mayaguez Part, Civil No. CS-88-896, for collection of money and damages; and a copy of a sworn statement of Nelson Torres Ruiz dated June 22, 1988.
.During the hearing the parties testified this suit was stayed by the filing of this petition.

\rule{\textwidth}{0.5pt}
Output a valid JSON object with the fields of {"label": [(one or more integers from 0-3 indicating the gap categories, expressed in a list)], "explanation": a short explanation justifying the label.}. Do not output anything else such as 'json' or newline characters or redundant spaces. Answer after output: 

output: 
\\\\
\textbf{Label}: [2]
\\\\
\textbf{Explanation}: The generated text provides detailed context and elaboration for each citation, whereas the target text chain cites them without additional detail. This indicates a target mismatch.
\end{minipage}
}
\caption{(2/2) A full example machine-generated legal analysis from \citep{abe2024clerc}, with previous context and cited sources provided Texts after the line break are example annotations.}
\label{fig:full_example2}
\end{figure*}

%% file: examples/example_3.tex
\begin{figure*}[t]
\centering
\fbox{
\footnotesize
\begin{minipage}{\textwidth}

\textbf{Generation}:\\
\dots
The court’s action was an error in law. In a recent case, the court held that Rule 25(a) (1) “is not designed to require a plaintiff to take affirmative action to locate the representative of the deceased party. Rather, it is designed to provide a mechanism for the court to manage the case in the event the deceased party has not been substituted.” 359 F.2d 292, 294 (C.A. 4, 1966). In another case, the court held that the Rule “does not require a plaintiff to institute machinery in order to produce a representative of the estate ad litem, pending appointment of the representative contemplated by law of the domicile of the deceased.” 379 F.2d 94, 96 (C.A. 7, 1967). In that case, the court noted that the “plaintiff’s attorney did not know whether probate of the will might be contested, or who would be appointed representative of the estate.” Id. at 96. The court held that the suggestion of death was ineffective to trigger the 90-day period under Rule 25(a) (1) where the suggestion did not identify a successor or representative of the deceased party. Id. at 96. See also, 4A Wright \& Miller, Federal Practice and Procedure: Civil 2d, 2451 (1969). The court’s action was an error in law. The judgment is reversed, and the case is remanded for further proceedings not inconsistent with this opinion. Reversed. (Emphasis added.) \\

\textbf{citations needed to make:} ['379 F.2d 94', '359 F.2d 292']
\\

\textbf{Target}:
\\
The amendment to Rule 25(a) (1) was intended to dispel unwarranted rigidity and allow more flexibility in substitution. “It was intended that liberal effect be given to the 1963 amendment.” Roscoe v. Roscoe, 126 U.S.App.D.C. 317, 322, 379 F.2d 94, 99 (1967). “[T]he 90 day period was not intended to act as a bar to otherwise meritorious actions.” Staggers v. Otto Gerdau Co., 359 F.2d 292, 296 (2d Cir. 1966). 

\textbf{reference\_case\_1:} 379 F.2d 94

the proper parties. \* \* \* Unless the motion for substitution is made not later than 90 days after the death is suggested upon the record by service of a statement of the fact of the death as provided herein for the service of the motion, the action shall be dismissed as to the deceased party.” Here not only had the 90-day period not expired when the court entered summary judgment, the appellant by her motion for reconsideration had specifically invoked the discretion of the court. Rule 6(b) provides pertinently that when “by these rules * * * an act is required or allowed to be done at or within a specified time, the court for cause shown may at any time in its discretion (1) with or without motion or notice order the period enlarged if request therefor is made before the expiration of the period originally prescribed * * Originally the Rule had precluded an extension of time for taking action under Rule 25(a) (1), but by purposeful amendment, it was sought to relieve against the hardship of the Court’s holding in Anderson v. Yungkau, 329 U.S. 482, 67 S.Ct. 428, 91 L.Ed. 436 (1947). It was intended that liberal effect be given to the 1963 amendment. Graham v. Pennsylvania Railroad, 119 U.S.App.D.C. 335, 342 F.2d 914 (1964), cert. denied, 381 U.S. 904, 85 S.Ct. 1446, 14 L.Ed.2d 286 (1965). We are constrained to reverse for further proceedings not inconsistent with this opinion. Reversed. The only “party” then 
\\
\textbf{reference\_case\_2:} 359 F.2d 292
\\
insertion of a “reasonable time” standard. In 1963, the Advisory Committee suggested the present rule and noted: “Present Rule 25(a) (1), together with present Rule 6(b), results in an inflexible requirement that an action be dismissed as to a deceased party if substitution is not carried out within a fixed period measured from the time of the death. The hardships and inequities of this unyielding requirement plainly appear from the cases. * * * The amended rulé establishes a time limit for the motion to substitute based not upon the time of the death, but rather upon the time information of the death is provided by means of a suggestion of death upon the record, i. e. service of a statement of the fact of the death.” See Notes of Advisory Committee on the Civil Rules, 28 U.S.C. Rule 25 (1964). Rule 6(b) of the Federal Rules of Civil Procedure was also amended in 1963 and the prohibition against extending the time for taking action under Rule 25 was eliminated. The Advisory Committee on the Civil Rules noted: “It is intended that the court shall have discretion to enlarge that period.” The amendments of Rules 6(b) and 25(a) (1) provided needed flexibility. It was assumed that discretionary extensions would be liberally granted. Movants under Rule 25 can ordinarily control when a death is “suggested upon the record” and appellants’ attorney was under no obligation to file his affidavit of Staggers’ death on the date he did. He could have filed 

\textbf{previous\_text:} 

LEVENTHAL, Circuit Judge: The District Court held that Rule 25(a) (1) of the Federal Rules of Civil Procedure required dismissal of the plaintiffs’ tort action because defendant’s counsel had filed a suggestion of death of the defendant yet plaintiff had not made any substitution of parties within 90 days. We reverse on the ground that the suggestion of death, which was neither filed by nor identified a successor or representative of the deceased, such as an executor or administrator, was ineffective to trigger the running of the 90-day period provided by the Rule. Mr. and Mrs. John Rende filed an action in the District Court individually and on behalf of their infant son who had been struck and injured by Alfred S. Kay while driving his car. On August 27, 1967, defendant Kay died. 

\dots

In our opinion the Rule, as amended, cannot fairly be construed, as the de fendant’s attorney argues, to make his suggestion of death operative to trigger the 90-day period even though he was neither a successor nor representative of the deceased, and gave no indication of what person was available to be named in substitution as a representative of the deceased. Counsel’s construction would open the door to a tactical maneuver to place upon the plaintiff the burden of locating the representative of the estate within 90 days. We can conceive of cases wherein even the lawyer retained to represent a defendant might know the defendant had died, yet not readily know where his estate would be administered. In the present case, plaintiff’s attorney did know the court of probate, but he did not know whether probate of the will might be contested, or who would be appointed representative of the estate. The tactic of the defendant’s attorney would place on plaintiff the burden, where no conventional representative was appointed for the estate in probate court, of instituting machinery in order to produce some representative of the estate ad litem, pending appointment of the representative contemplated by law of the domicile of the deceased.

(GAP CATEGORIES AND EXPLANATION INCLUDED NEXT PAGE)
\end{minipage}
}
\caption{(1/2) An example machine-generated legal analysis which contains \textit{redundancy}, \textit{structural mismatch}, \textit{claim hallucination}, \textit{citation hallucination}, and \textit{retrieval inaccuracy}.}
\label{fig:example_3}
\end{figure*}

\begin{figure*}[t]
\centering
\fbox{
\footnotesize
\begin{minipage}{\textwidth}
(CONTINUED FROM LAST PAGE)\\\\
\textbf{Gap Categories:}\\
Redundancy ($G^5$), Structural Mismatch ($G^8$), Claim Hallucination ($G^9$), Citation Hallucination ($G^{10}$), and Retrieval Inaccuracy ($G^{11}$).\\\\
\textbf{Explanation:}\\
1. The generation has redundancy because it repeats "The court's action was an error in law" verbatim twice and also repeats "Reversed" multiple times. 2. The generation has structural mismatch because it prematurely concludes the case with "Reversed", which is structurally inappropriate. 3. It contains claim hallucination, as when it elaborates case 379 F.2d 94, it should focus on the aspects related to "It was
intended that liberal effect be given to the 1963 amendment." as this is the relevant claim. However, it talks about "does not require a plaintiff to institute machinery
in order to produce a representative of the estate ad litem, pending appointment of the representative contemplated by
law of the domicile of the deceased", which is rather irrelevant. 4. The generation also suffers from retrieval inaccuracy. The generation should retrieve the salient chunk from case 359
F.2d 292 centering on "T]he 90 day period was not intended to act as a bar to otherwise meritorious actions.", yet the retrieved chunk is not so relevant.
\end{minipage}
}
\caption{(2/2) Explanation of the example in Figure \ref{fig:example_3}.}
\label{fig:example_3_exp}
\end{figure*}

%% file: examples/example_11.tex
\begin{figure*}[t]
\centering
\fbox{
\footnotesize
\begin{minipage}{\textwidth}
\textbf{Generation}:\\
\dots
This case presents the question whether an employee welfare benefits plan creates an entitlement to lifetime benefits or just to benefits that can be terminated by an amendment to the plan. The plan administrator, Motorola, amended the plan to place a two-year limit on benefits for disability resulting from certain mental conditions, including the plaintiff's condition. The plaintiff, Marrs, argues that the plan's provision stating that no amendment shall adversely affect the rights of any participant to receive benefits with respect to periods of disability prior to the adoption date of the amendment is violated. He interprets "periods of disability prior to the adoption date" to mean one or more periods of disability that began before the plan was amended but may not have ended before then. However, the plan administrator's interpretation is reasonable, and we are inclined to stop with that observation. The Supreme Court's decision in Glenn v. Metropolitan Life Ins. Co. (128 S.Ct. 2343, 171 L.Ed.2d 299 (2008)) is relevant in this case. The Court held that a conflict of interest by the plan administrator should be given weight in judicial review of the denial of benefits. However, the weight to be given to the conflict of interest depends on the circumstances of the case. In this case, the plan administrator had an incentive to resolve a close case in favor of a denial of benefits, but this incentive may be outweighed by other incentives, such as an employer's interest in maintaining a reputation among current and prospective employees for fair dealing. The employees who actually decide benefits claims at the plan- administrator level may not be acutely concerned with the financial implications of a benefits award for their employer. We conclude that the plan administrator's decision to deny benefits is reasonable and that the conflict of interest does not render the decision unreasonable. The judgment of the district court is Affirmed. \underline{ References: 1. Glenn v. Metropolitan Life Ins. Co., 128 S.Ct. 2343, 171 L.Ed.2d 299 (2008) 2. Marrs v. Motorola,} \underline{ 908 F.2d 1385 (7th Cir. 1990) 3. Rogers v. Department of Health \& Environmental Control, 174 F.3d 431, 435 (4th Cir. 1999)} \underline{ 4. Kahane v. UNUM Life Ins. Co., 563 F.3d 1210, 1212 (11th Cir.} \\

\textbf{citations needed to make:} ['564 F.3d 856', '576 F.3d 240', '560 F.3d 620', '504 F.3d 665', '908 F.2d 1385', '855 F.2d 394']\\

\textbf{Target}:
\\
There are two ways to read the majority opinion. One, which tracks its language and has been echoed in opinions in this and other circuits, e.g., Jenkins v. Price Waterhouse Long Term Disability Plan, 564 F.3d 856, 861-62 (7th Cir.2009); Holland v. Int’l Paper Co. Retirement Plan, 576 F.3d 240, 246-49 (5th Cir.2009), makes the existence of a conflict of interest one factor out of many in determining reasonableness. That sounds like a balancing test in which unweighted factors mysteriously are weighed. Such a test is not conducive to providing guidance to courts or plan administrators. “Multifactor tests with no weight assigned to any factor are bad enough from the standpoint of providing an objective basis for a judicial decision; multifactor tests when none of the factors is concrete are worse.” Menard, Inc. v. Commissioner, 560 F.3d 620, 622-23 (7th Cir.2009) (citations omitted); see also Sullivan v. William A. Randolph, Inc., 504 F.3d 665, 671 (7th Cir.2007); Short v. Belleville Shoe Mfg. Co., 908 F.2d 1385, 1394 (7th Cir.1990) (concurring opin ion); Stevens v. Tillman, 855 F.2d 394, 399-400 (7th Cir.1988). \\

\textbf{previous\_text:} 
\\
POSNER, Circuit Judge. This suit under ERISA for disability payments presents the recurring question whether an employee welfare benefits plan creates an entitlement to lifetime benefits rather than just to benefits that can be terminated by an amendment to the plan. In 1997 Michael Marrs, an employee of Motorola, ceased working because of a psychiatric condition and began drawing disability benefits under Motorola’s Disability Income Plan. Six years later Motorola amended the plan to place a two-year limit on benefits for disability resulting from certain “Mental, Nervous, Alcohol, [or] Drug-Related” (MNAD) conditions, including Marrs’s. Such limitations on MNAD conditions are common in employee disability plans.

Then too, the employees who actually decide benefits claims at the plan-administrator level may not be acutely concerned with the financial implications of a benefits award for their employer. Id. at 821; Perlman v. Swiss Bank Corp. Comprehen sive Disability Protection Plan, 195 F.3d 975, 981 (7th Cir.1999). But especially when a firm is struggling (which may or may not be the case here — there is nothing in the record bearing on the question), an opportunity for short-run economies may dominate decision making by benefits officers. In any event, a majority of the Supreme Court Justices consider the potential conflict of interest of a plan administrator (or its staff) serious enough to be given weight in judicial review of the denial of benefits. But how much weight should it be given? The nub of the Glenn opinion is the following passage: [W]hen judges review the lawfulness of benefit denials, they will often take account of several different considerations of which a conflict of interest is one. This kind of review is no stranger to the judicial system. Not only trust law, but also administrative law, can ask judges to determine lawfulness by taking account of several different, often case-specific, factors, reaching a result by weighing all together. In such instances, any one factor will act as a tiebreaker when the other factors are closely balanced, the degree of closeness necessary depending upon the tiebreaking factor’s inherent or case-specific importance. The conflict of interest at issue here, for example, should prove more important (perhaps of great importance) where circumstances suggest a higher likelihood that it affected the benefits decision, including, but not limited to, cases where an insurance company administrator has a history of biased claims administration. It should prove less important (perhaps to the vanishing point) where the administrator has taken active steps to reduce potential bias and to promote accuracy, for example, by walling off claims administrators from those interested in firm finances, or by imposing management checks that penalize inaccurate decisionmaking irrespective of whom the inaccuracy benefits. 128 S.Ct. at 2351 (citations omitted). A dissent by Justice Scalia argued that a conflict of interest should prompt an inquiry only into the existence of improper motive that would render the plan administrator’s decision unreasonable. If the decision is reasonable, he argued, in the sense in which “a reasonable decision is one over which reasonable minds seeking the ‘best’ or ‘right’ answer could disagree,” the fact that the administrator had a conflict of interest is irrelevant, id. at 2360, “unless the conflict actually and improperly motivates the decision.” Id. at 2357 (emphasis in original).
\\\\
    (GAP CATEGORIES AND EXPLANATION ARE INCLUDED IN THE NEXT PAGE).
    \end{minipage}
}
    \caption{(1/2) An example machine-generated legal analysis that contains \textit{citation format mismatch}, \textit{structural mismatch}, \textit{citation hallucination}, and \textit{stylistic mismatch}.}
\label{fig:example_11}
\end{figure*}

\begin{figure*}
    \centering
    \fbox{
    \begin{minipage}{\textwidth}
    \footnotesize
        (CONTINUED FROM LAST PAGE)\\\\     
        \textbf{Gap Categories:}\\
Citation Format Mismatch ($G^6$), Stylistic Mismatch ($G^7$), Structural Mismatch ($G^8$), and Citation Hallucination ($G^{10}$).
\\\\
\textbf{Explanation:}\\
1. The generation has citation format mismatch because it includes inappropriate citation formats (underlined) to legal writing. 2. The generation has stylistic mismatch because it should not use language like "This case" when it should be continuing the generation, and using a third-person pronoun makes it appear like analyzing a separate piece of text rather than continuing it. 3. The generation has structural mismatch, since it "conclude(s)" the case, which is inappropriate structurally. 4. The generation contains citation hallucination as it fails to cite most of the citations needed to make.
    \end{minipage}
    }
    \caption{(2/2) Explanation of the example in Figure \ref{fig:example_11}.}
\label{fig:example_11_exp}
\end{figure*}

%% file: examples/target_mismatch_examples.tex
\begin{figure*}[t]
\centering
\fbox{
\footnotesize
\begin{minipage}{\textwidth}

\textbf{Generation:}\\
CONCLUSIONS OF LAW (continued) In Butner v. United States, 440 U.S. 48, 55, the Supreme Court emphasized that property rights in the assets of a bankrupt's estate are determined by state law unless a federal interest dictates otherwise. In this case, we must determine the debtor's obligations under Puerto Rican law and assess the applicability of interest on the debt owed to Nelson Torres Ruiz. Consistent with the principles outlined in 114 B.R. 326, the determination of whether the debtor has a legal or equitable interest in the obligation to return the \$3,000 option money necessitates an analysis grounded in applicable nonbankruptcy state law. Article 1061 of the P.R. Civil Code, 31 Laws of P.R.Ann., Section 3025, establishes that where a debtor fails to perform an obligation, the creditor is entitled to interest as indemnification for damages and losses incurred. Debtor's claim of impossibility of performance is addressed under Article 1138 of the Civil Code of Puerto Rico, but as established earlier, such a defense is unwarranted when the impossibility is not due to legal or physical barriers (31 Laws of P.R.Ann., Section 3193). Given the absence of such barriers, the return of the \$3,000 is mandated. Furthermore, in alignment with 117 B.R. 15, the initiation of bankruptcy proceedings does not alter the creditor’s right to claim interest at the legal rate. Should the contract fail to specify an interest rate, Article 1061 defaults to a legal interest rate of 6\% per annum. Thus, Mr. Bonilla is obliged to reimburse the full amount of \$3,000 plus the applicable legal interest rate of 6\% from the expiration of the option contract until the filing of the bankruptcy petition.
\\\\
\textbf{Target:}\\
See generally, 3 Collier On Bankruptcy, (15th ed.) paragraph 502.02 (1990). See also, Butner v. U.S., 440 U.S. 48, 99 S.Ct. 914, 59 L.Ed.2d 136 (1979); In re MacDonald, 114 B.R. 326 (D.Mass. 1990); In re Milford Common J.V. Trust, 117 B.R. 15 (Bkrtcy.Mass., 1990).
\\\\
\textbf{Gap Categories:}\\
Chain vs. Parallel ($G^{12}$) \\

\textbf{Explanation:}\\
The target text does not elaborate any of the cases cited whereas the generation chooses to elaborate each citation.
\end{minipage}
}

\caption{An example machine-generated legal analysis which contains \textit{chain vs. parallel}.}
\label{fig:chain_cite}
\end{figure*}

\begin{figure*}[t]
\centering
\fbox{
\footnotesize
\begin{minipage}{\textwidth}

\textbf{Generation:}\\
In Reference case 47 M.J. 370, the court held that the automatic total-forfeiture rule and the 14-day provision of Article 57(a)(1) violate the Ex Post Facto Clause. \underline{Similarly}, in Reference case 45 M.J. 567, the court found that the Article 57 amendment does not violate the ex post facto prohibition, as it does not increase the duration of the punishment.
\\\\
\textbf{Target:}\\
Appellant’s ex post facto arguments were resolved by the United States Court of Appeals for the Armed Forces in United States v. Gorsky 47 M.J. 370 (1997). We intend to apply that decision despite appellate government counsel’s argument that we should ignore our superior Court’s opinion and adhere to our decision in United States v. Pedrazoli, 45 M.J. 567(A.F.Ct.Crim.App.1997), \textbf{which Gorski essentially reversed}. See United States v. Plumb, 47 M.J. 771 (A.F.Ct.Crim. App.1997).
\\\\
\textbf{Gap Categories:}\\
Agree vs. Disagree ($G^{13}$) and Citation Hallucination ($G^{10}$)  \\

\textbf{Explanation:}\\
In the target text, case 47 M.J. 370 reversed the ruling of case 45 M.J. 567 (highlighted in bold). However, the generation illustrates the two cases in a parallel manner, as shown by the conjuctive adverb "\underline{Similarly}". Moreover, the generation also does not mention case 47 M.J. 771, which is cited in the target.
\end{minipage}
}

\caption{An example machine-generated legal analysis which contains \textit{agree vs. disagree} and \textit{citation hallucination}.}
\label{fig:reverse_cite}
\end{figure*}

\begin{figure*}[t]
\centering
\fbox{
\footnotesize
\begin{minipage}{\textwidth}

\textbf{Generation:}\\
This dismissal might be an error. According to established precedents, an action can be dismissed as frivolous under 28 U.S.C. § 1915(d) only if it is beyond doubt that the petitioner can prove no set of facts in support of his claim that would entitle him to relief (699 F.2d 434; 741 F.2d 209).
\\\\
\textbf{Target:}\\
A district court may dismiss an action as frivolous only if it appears beyond a doubt that the plaintiff can prove no set of facts in support of his claim which would entitle him to relief. Smith v. Bacon, 699 F.2d 434, 436 (8th Cir.1983). A complaint which is good against a motion to dismiss for failure to state a claim may not be dismissed as frivolous under 28 U.S.C. § 1915(d). Horsey v. Asher, 741 F.2d 209, 212 (8th Cir.1984).
\\\\
\textbf{Gap Categories:}\\
Compound Cite ($G^{14}$)  \\

\textbf{Explanation:}\\
The target text cites two cases to illustrate two claims ("A district court may dismiss an action as frivolous only if it appears beyond a doubt that the plaintiff can prove no set of facts in
support of his claim which would entitle him to relief." and "A complaint which
is good against a motion to dismiss for failure to state a claim may not be dismissed as frivolous under 28 U.S.C. § 1915(d).") However, the generation combines the two claims ("an action can be dismissed as frivolous under 28 U.S.C. § 1915(d) only if it is beyond doubt that the petitioner can prove no set of facts in support of his claim that would entitle him to
relief ") and cites the two cases together.
\end{minipage}
}

\caption{An example machine-generated legal analysis which contains \textit{compound cite}.}
\label{fig:compound_cite}
\end{figure*}

%% file: examples/annotation_instructions.tex
\begin{figure*}[t]
    \centering
    \fbox{
    \footnotesize
    \begin{minipage}{\textwidth}
    \underline{Instructions for annotators:}
    \\\\
\textbf{Task Overview}

You are tasked to classify categories of gaps between machine-generated and human-written legal analysis. 

\textbf{Definitions:}

\textbf{generation}: machine-generated legal analysis.

\textbf{target}: human-written legal analysis. Note that the target is only one form of acceptable legal analysis. There are other acceptable legal analysis. It is possible for a generation to not match with the target but still considered acceptable.
\newline
\textbf{previous\_context}: we set the goal of LLM to generate a paragraph of legal analysis and feed in the previous context to this paragraph as the input.

\textbf{cited\_paragraphs}: in addition to the previous context, we also feed in the other paragraphs that are supposed to be cited in this generation.

\textbf{citation}: citation refers to the special string which points to a legal case, with style and format specified by the Bluebook.

\textbf{claim}: the sentence which is supported by the citation, i.e. the case referred to. Claim usually appears in the vicinity of the citation.

\textbf{Intrinsic Gaps}: the presence of intrinsic gaps signals that the machine-generated legal analysis is an unacceptable form. We can tell intrinsic gaps exist by \textit{only} looking at the previous context and the generation itself.

\textbf{Extrinsic Gaps}: extrinsic gaps, as its name suggests, can be discovered by comparing the generation with external texts, i.e. the cited paragraphs or the target paragraph that can be seen as the "answer". Extrinsic gaps contain two kinds: citation content mismatch and target mismatch. Target mismatch does not indicate that the generated legal analysis is necessarily wrong.
\\\\
\textbf{Annotation Instructions:}
\\
Receiving the following prompt, a language model will generate a paragraph of legal analysis, but often times they make different kinds of errors and mismatches. 
\\\\
\texttt{
\textbf{User prompt}: \\
Here are some reference articles for legal cases: \\
\# Reference case \{case\_key\_1\} \\
\{text of cited case 1\} \\
\# Reference case \{case\_key\_2\} \\
\{text of cited case 2\} \\
$\dots$ \\
\# Reference case \{case\_key\_N\} \\
\{text of cited case N\} \\\\
Here is the text I've written so far: \\
\# Paragrah \\
\{previous\_text\} \\\\
Continue to write it following the style of my writeup. Your answer contains 100 to 400 words. You must explicitly use the reference cases and mention their reference ids, i.e. \{case\_key\_1\}, \{case\_key\_2\} \dots \{case\_key\_N\}. Wrap your answer with <answer></answer>. Make your answer concise and avoid redundant languages. \\
}

The instructions for you to classify these errors and mismatches are as follows:
\\\\
1. \textbf{Intrinsic gap}:\\
This category refers to generation that is unacceptable, due to the language model has fundamentally failed to follow the instruction, or make a lot of redundancy, or generate something that does not look like legal text (structural mismatch). More specificially, if it makes one or more of the following:
\begin{itemize}
    \item Redundancy (sentence-level, appearing as neural degeneration): the generation appears to make repetitive statements that do not add more meaning to the analysis. For example, multiple occurences of an exact sentence or phrase.
    \item Citation Format Mismatch: the generation appears not matching with the citation format of the standard Bluebook.
    \begin{itemize}
        \item Please be aware that, for example, 440 U.S. 48, 55' is a proper format. Although its full citation should be 'Butner v. United States, 440 U.S. 48, 55 (1979)', the format '440 U.S. 48, 55' is still acceptable as a concise form.
    \end{itemize}
    \item Structural Mismatch: the generation appears to generate the document from scratch (like containing words such as "ORDER" which only appear in the beginning).
    \item Stylistic Mismatch: contain sentences that do not match the styles of legalese.
\end{itemize}

\textbf{If this type of gaps is present, add the label `1`. Continue to item 2.}

Side note: You should be able to classify this purely based on the generation itself, without having to look at cited examples.
\\\\
2.\textbf{Target mismatch}:

While language model's generated text may be obviously wrong and substantively different from the target (i.e. the original/target text from the case), the claims it makes are still logically and factually sound and can be seen as acceptable. This could be because
\\\\
(CONTINUED IN NEXT PAGE)
    \end{minipage}
    }
    \vspace{-2mm}
    \caption{(1/2) Annotation instructions for human annotators.}
    \label{fig:instructions-fig1}
\end{figure*}

\begin{figure*}
    \centering
    \fbox{
    \footnotesize
    \begin{minipage}{\textwidth}
    (CONTINUED FROM LAST PAGE)
    \begin{itemize}
    \item Chain cite: the citations appear in a chain cite but the generation cites them parallely, or the other way around.
    \begin{itemize}
        \item Clarification: "The rule that certain acts of a creditor in the course of a bankruptcy proceeding and during the statutory period for filing proof of claim, may give rise to something equivalent to a proof of claim and afford a sufficient basis for allowing an amendment after the statutory period for filing, was recognized and applied in many cases decided before the 1938 amendment of the Bankruptcy Act. See In re Atlantic Gulf \& Pacific S. S. Corporation, D.C., 26 F.2d 751; In re Fant, D.C., 21 F.2d 182; Globe Indemnity Co. of Newark, N. J., v. Keeble, 4 Cir., 20 F.2d 84; In re Coleman \& Titus Corporation, D.C., 286 F. 303; In re Roeber, 2 Cir., 127 F. 122." \textbf{would be a chain cite} because all of these citations support the previous claim "The rule that certain acts of a creditor in the course of a bankruptcy proceeding and during the statutory period for filing proof of claim, may give rise to something equivalent to a proof of claim and afford a sufficient basis for allowing an amendment after the statutory period for filing, was recognized and applied in many cases decided before the 1938 amendment of the Bankruptcy Act."
    \end{itemize}
    \item Agree versus disagree: the citations reverse the ruling in each other but the generation cites them parallely, or the other way around.
    \item Compound cite: the citations of different cases are cited together, separated by semicolons, or the other way around.
\end{itemize}
Although it does not match with the target, it is still considered somewhat acceptable, but we should label it out.

If this type of mismatch is present add the label "2". Continue to item 3.
\\\\
3. \textbf{Citation Mismatch}: \\
The language model's generated text does not align with the content of the citation it points to. This might be because one or more of the following:
\begin{itemize}
    \item Claim Hallucination: the claim supported by the citation is not truthful or not related to the context or from cited paragraphs or the previous context. The generated text makes different and possibly (although not necessarily) contradictory claims about one or more citations, which you can check from comparing to the reference case. Or, the generated text attributes information from one citation to a different citation.
\item Retrieval Inaccuracy: the claims supported by the citation is not relevant because the cited paragraph looks irrelevant compared to the target paragraph. 
\item Citation Hallucination: the citation is non-existent or pulled from a citation in the cited paragraphs or the previous context, or there misses a citation (the generated text fails to use one of the citations that were given to it).
\end{itemize}

If this type of mismatch is \textbf{present}, add the label "3" and move on to the next example. If none of the above errors are present, label "0".\\

Note that where an example falls into multiple categories, you should include both labels, separated by a comma.
    \end{minipage}
}
    \caption{(2/2) Annotation instructions for human annotators.}
    \label{fig:instructions-fig2}
\end{figure*}

%% file: examples/prompts_to_detectors.tex
\begin{figure*}[t]
\centering
\fbox{
\footnotesize
\begin{minipage}{\textwidth}

\textbf{System Prompt:}\\
You are a trained lawyer from Silicon Valley with a computer science background. Now, you are asked to annotate legal analysis generated by large language models and classify the errors and mismatch made by these models. To produce these legal analysis, a language model will receive the following prompt:\\

Here are some reference articles for legal cases:\\ 
\# Reference case\\ \{case\_key\_1\}
\{text of cited case 1\} \\
\# Reference case\\ \{case\_key\_2\}
\{text of cited case 2\}\\
...\\
\# Reference case\\ {case\_key\_N} \{text of cited case N\}\\\\
Here is the text I’ve written so far:\\ \# Paragraph
\{previous\_text\}
Continue to write it following the style of my writeup. Your answer contains 100 to 400 words. You must explicitly use the reference cases and mention their reference ids, i.e. \{case\_key\_1\}, \{case\_key\_2\} . . . \{case\_key\_N\}. Wrap your answer with <answer></answer>. Make your answer concise and avoid redundant languages.

Receiving the prompt above, a language model will generate a paragraph of legal analysis, but often times they make different kinds of errors and mismatches. 

The instructions for you to classify these errors and mismatches are as follows:

You should classify the LLM-generated legal analysis to these categories:

\{\texttt{Summary of the gap categories, same from the instructions to human annotators.}\}

Here are some examples for demonstration:

\{\texttt{Example annotation 1}\}

\{\texttt{Example annotation 2}\}

\vdots

\{\texttt{Example annotation k}\}

-----------------------------------------------------------------------------------------------

--End Demonstration--

Now, we will give you more instances and have you annotate 1, 2, 3, or 0. Output a json object containing the label and explanation for each example. If you label a 3, please elaborate the explanation for it a bit more.
\rule{\textwidth}{0.5pt}
\textbf{User Prompt}:\\
\text{Generation:}
\{\texttt{generation}\}\\\\
\text{citations needed to make:}
\{\texttt{[citation\_1, \dots, citation\_N]}\}\\\\
\text{Target:}\{\texttt{target}\}\\\\
\text{reference\_case\_1}: \{\texttt{case\_key\_1}\}\\\{\texttt{reference\_case\_1}\}\\
\vdots\\
\text{reference\_case\_N}: \{\texttt{case\_key\_N}\}\\
\{\texttt{reference\_case\_N}\}\\\\
\text{previous\_text:} \{\texttt{previous\_text}\}
\\\\
Output a valid JSON object with the fields of {"label": [(one or more integers from 0-3 indicating the gap categories, expressed in a list)], "explanation": a short explanation justifying the label.}. Do not output anything else such as 'json' or newline characters or redundant spaces. Answer after output: 

output: 

\end{minipage}
}

\caption{Prompts to LLM-based detectors. The number of $k$ varies from \{4, 8, 16, 20\} in our ablation studies.}
\label{fig:prompts}
\end{figure*}